\pdfoutput=1
\documentclass[11pt]{article}
\usepackage[preprint]{acl}
\usepackage{times}
\usepackage{latexsym}
\usepackage[T1]{fontenc}
\usepackage[utf8]{inputenc}
\usepackage{microtype}
\usepackage{inconsolata}
\usepackage{graphicx}
\usepackage{booktabs}
\usepackage{multirow}
\usepackage{xcolor}
\usepackage{pgfplots}
\usepackage{pgfplotstable}
\pgfplotsset{compat=1.18}
\usetikzlibrary{positioning,arrows.meta,fit,backgrounds}
\usepackage{enumitem}
\usepackage{float}
\usepackage{amsmath,amssymb}
\setlist{nosep,leftmargin=*}
\newcommand{\MainSense}{97.8}
\newcommand{\MainSenseLo}{92.5}
\newcommand{\MainSenseHi}{99.4}
\newcommand{\MainSenseK}{91}
\newcommand{\MainSenseN}{93}
\newcommand{\MainPair}{97.0}

\newcommand{\MainPairK}{131}
\newcommand{\MainPairN}{135}
\newcommand{\MainCorrect}{96.2}
\newcommand{\MainCorrectLo}{91.4}
\newcommand{\MainCorrectHi}{98.4}

\newcommand{\MainSchema}{100.0}

\newcommand{\MainDiagram}{99.2}

\newcommand{\MainDiagramK}{130}
\newcommand{\MainDiagramN}{131}
\newcommand{\MainLogic}{4.90}

\newcommand{\MainClauseSilly}{2}

\newcommand{\MainFirstMed}{249}
\newcommand{\MainFullMed}{1018}
\newcommand{\MainFirstNinety}{300}
\newcommand{\MainFullNinety}{1185}
\newcommand{\MainFullMax}{2090}
\newcommand{\MainRetries}{1}

\newcommand{\AltSense}{98.9}
\newcommand{\AltSenseLo}{94.2}
\newcommand{\AltSenseHi}{99.8}

\newcommand{\AltPair}{98.5}

\newcommand{\AltCorrect}{96.2}
\newcommand{\AltCorrectLo}{91.4}
\newcommand{\AltCorrectHi}{98.4}

\newcommand{\AltLogic}{4.92}

\newcommand{\AltClauseSilly}{1}

\newcommand{\VoneSense}{96.4}

\newcommand{\VonePair}{89.7}

\newcommand{\VonePairK}{104}

\newcommand{\VoneCorrect}{94.2}

\newcommand{\VoneClauseSilly}{29}

\newcommand{\VtwoSense}{98.8}

\newcommand{\VtwoPair}{98.3}

\newcommand{\VtwoPairK}{114}

\newcommand{\VtwoCorrect}{95.8}

\newcommand{\VtwoClauseSilly}{1}

\newcommand{\VthreeSense}{96.4}

\newcommand{\VthreePair}{96.6}

\newcommand{\VthreeCorrect}{94.2}

\newcommand{\VthreeClauseSilly}{2}

\newcommand{\JudgeSameAnswers}{131}
\newcommand{\JudgeVerdictsChanged}{24}
\newcommand{\JudgeScoreChanged}{17}
\newcommand{\JudgeScoreMaxStep}{1}
\newcommand{\JudgeClauseChanged}{9}
\newcommand{\JudgeSenseFlips}{2}

\newcommand{\JudgeSameConfigAgree}{120}

\newcommand{\MainCorrectStrict}{94.7}
\newcommand{\MainCorrectStrictLo}{89.4}
\newcommand{\MainCorrectStrictHi}{97.4}
\newcommand{\MainCorrectStrictK}{124}
\newcommand{\MainCorrectStrictN}{131}
\newcommand{\SetCases}{131}
\newcommand{\SetDocs}{23}
\newcommand{\SetPapers}{16}
\newcommand{\SetFields}{8}
\newcommand{\SetKindPolysemous}{93}
\newcommand{\SetKindTerm}{30}
\newcommand{\SetKindPassage}{8}
\newcommand{\SetNonPdf}{32}
\newcommand{\SetMultiFieldTerms}{16}
\newcommand{\FmtDocx}{9}
\newcommand{\FmtEpub}{6}
\newcommand{\FmtMd}{8}
\newcommand{\FmtPdf}{99}
\newcommand{\FmtPptx}{5}
\newcommand{\FmtTxt}{4}
\newcommand{\FigCases}{28}
\newcommand{\FigDocs}{14}

\newcommand{\FigCorrectLo}{87.9}
\newcommand{\FigCorrectHi}{100.0}
\newcommand{\FigCorrectK}{28}
\newcommand{\FigCorrectN}{28}

\newcommand{\FigGroundedLo}{87.9}
\newcommand{\FigGroundedHi}{100.0}
\newcommand{\FigGroundedK}{28}
\newcommand{\FigGroundedN}{28}

\newcommand{\FigInventedLo}{0.6}
\newcommand{\FigInventedHi}{17.7}
\newcommand{\FigInventedK}{1}
\newcommand{\FigInventedN}{28}

\newcommand{\FigSchemaK}{28}
\newcommand{\FigSchemaN}{28}
\newcommand{\FigScoreFive}{26}
\newcommand{\FigMedMs}{882}
\newcommand{\FigMaxMs}{4420}

\newcommand{\pct}[1]{#1\,\%}
\newcommand{\ci}[2]{[#1, #2]}
\newcommand{\aynaz}{\textsc{Aynaz}}
\definecolor{hl}{HTML}{E5C443}
\definecolor{accent}{HTML}{24408E}
\definecolor{muted}{HTML}{8A93A3}

\title{Aynaz: Offline, Document-Grounded Explanations of Words and Figures\\with Small Local Language Models}

\author{Azharul Islam \\
  Hanyang University, Seoul, Republic of Korea \\
  \texttt{azharul001@hanyang.ac.kr}}

\begin{document}
\maketitle

\begin{abstract}
Readers of technical documents often meet words whose meaning depends on the field (\emph{kernel}, \emph{moment} or \emph{resistance} mean different things in a physics paper, a statistics paper and a clinical study) and figures that are hard to read.
A common alternative is to upload the document to a cloud chat assistant, at the cost of network access, usage quota and, for confidential documents, privacy.
We present \aynaz{}, a free desktop reader that explains a selected word the way the surrounding document uses it, with a meaning, an example and a small diagram, and explains a figure the reader draws a box around.
Everything runs locally: Qwen3-4B for text and Qwen3-VL-4B for figures, both 4-bit (Q4\_K\_M) quantised and served by llama.cpp, with a compact context packet, BM25 retrieval and schema-constrained decoding.
On \SetCases{} cases built around words with several meanings across \SetFields{} fields, scored by a larger model, \aynaz{} picks the document's sense in \MainSenseK{} of \MainSenseN{} polysemous cases (\pct{\MainSense}, 95\,\% CI \ci{\MainSenseLo}{\MainSenseHi}); its first words appear after a median of \MainFirstMed{}\,ms and the full answer after \MainFullMed{}\,ms on one consumer GPU.
All \FigCases{} figure explanations were judged correct in context (95\,\% CI lower bound \pct{\FigCorrectLo}).
In two evaluation runs the answers were identical, while the judge's verdicts moved in \JudgeVerdictsChanged{} of \SetCases{} cases when its server configuration changed.
\end{abstract}

\begin{figure*}[t]
  \centering
  \includegraphics[width=0.88\textwidth, trim=200 95 49 40, clip]{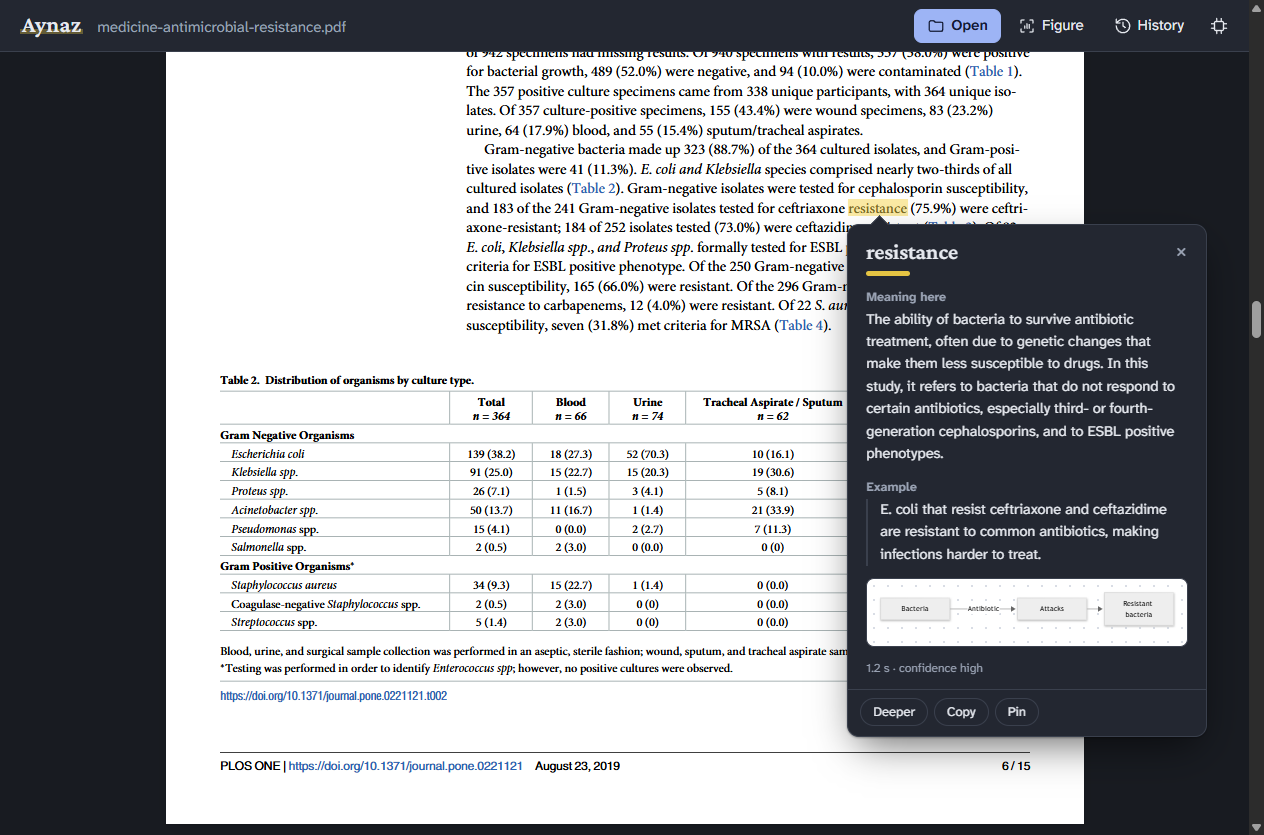}
  \caption{\aynaz{} explaining the highlighted word \emph{resistance} in an open-access clinical study: the meaning as used in this document, an example and a diagram. Captured from the application (version 0.5.4); an illustration, not one of the evaluation answers.}
  \label{fig:teaser}
\end{figure*}

\section{Introduction}

A word such as \emph{resistance} names antibiotic survival in a clinical study, opposition to current in a circuit and a material's ability to withstand deformation in a materials paper.
A reader who meets an unfamiliar use needs the sense \emph{this} document intends, not a dictionary's list of senses, which is word sense disambiguation in its most practical form \citep{navigli2009wsd}.
Figures raise the same problem in a harder form: a reader may see the axes and curves of a chart without knowing what the authors want it to show.

Large language models can explain both, and readers can copy passages, or upload whole documents, into a cloud chat assistant.
That route has costs that are easy to overlook while reading: the whole document leaves the reader's computer, every question uses network access and usage quota, and documents that are confidential (patient records, a manuscript under review, a company report) often may not be uploaded at all.
It also takes the reader out of the document and into a separate conversation.

\aynaz{} keeps the explanation in place and on the reader's machine.
The reader selects a word or phrase; a popup at the selection gives its meaning \emph{as used in this document}, a concrete example and a small diagram (Figure~\ref{fig:teaser}).
For a figure, the reader draws a box around it and \aynaz{} says what it shows, how to read it and the takeaway.
The models are small open-weight models, Qwen3-4B for text \citep{yang2025qwen3} and Qwen3-VL-4B for figures \citep{qwen2025qwen3vl}, served locally by llama.cpp \citep{llamacpp}, so the application works offline after a one-time model download.

The central question is whether a 4-bit (Q4\_K\_M) quantised 4B model, which one consumer GPU (an RTX 5070 Ti) ran in about a second per answer, picks the document's sense reliably enough to be useful.
Our contributions are:
\begin{itemize}
  \item a document-grounded explanation pipeline sized for small local models: a context packet of about 2k tokens, schema-constrained decoding, diagram validation with one retry, and streaming of the meaning as it is generated (\S\ref{sec:system});
  \item an evaluation built around words with several meanings, including \emph{cross-field pairs} that require the same term to be explained correctly in two fields' documents, and a separate figure evaluation (\S\ref{sec:eval});
  \item results on a consumer GPU for sense accuracy, explanation quality and latency, a prompt ablation, an engine and quantisation comparison, and an analysis of how much the model judge alone moves the headline numbers (\S\ref{sec:results});
  \item \aynaz{} itself, a free application for Windows, macOS (Apple silicon) and Linux.
\end{itemize}

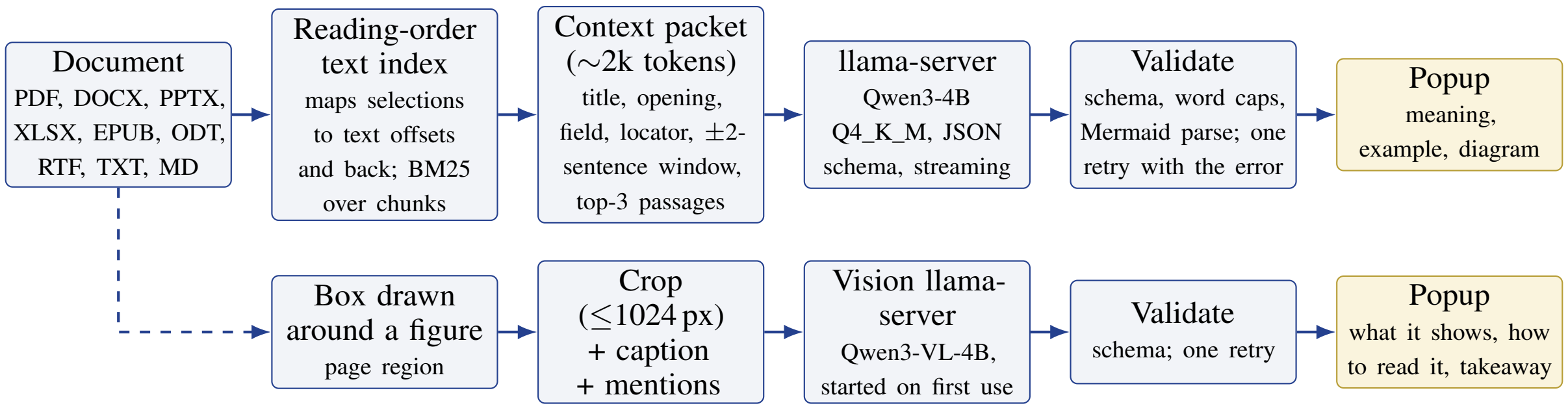
\begin{figure*}[t]
\centering
\begin{tikzpicture}[
  box/.style={draw=accent, rounded corners=2pt, fill=accent!6, align=center, font=\footnotesize, minimum height=1.05cm, text width=2.12cm, inner sep=2.5pt},
  result/.style={box, draw=hl!80!black, fill=hl!18},
  arr/.style={-{Latex[length=2mm]}, thick, accent},
  lab/.style={font=\scriptsize\itshape, text=black!70}]
  \node[box] (doc) {Document\\{\scriptsize PDF, DOCX, PPTX, XLSX, EPUB, ODT, RTF, TXT, MD}};
  \node[box, right=0.4cm of doc] (idx) {Reading-order text index\\{\scriptsize maps selections to text offsets and back; BM25 over chunks}};
  \node[box, right=0.4cm of idx] (pkt) {Context packet (${\sim}$2k tokens)\\{\scriptsize title, opening, field, locator, $\pm$2-sentence window, top-3 passages}};
  \node[box, right=0.4cm of pkt] (llm) {llama-server\\{\scriptsize Qwen3-4B Q4\_K\_M, JSON schema, streaming}};
  \node[box, right=0.4cm of llm] (val) {Validate\\{\scriptsize schema, word caps, Mermaid parse; one retry with the error}};
  \node[result, right=0.4cm of val] (pop) {Popup\\{\scriptsize meaning, example, diagram}};
  \draw[arr] (doc) -- (idx); \draw[arr] (idx) -- (pkt); \draw[arr] (pkt) -- (llm); \draw[arr] (llm) -- (val); \draw[arr] (val) -- (pop);
  \node[box, below=0.55cm of idx] (boxsel) {Box drawn around a figure\\{\scriptsize page region}};
  \node[box, right=0.4cm of boxsel] (crop) {Crop ($\le$1024\,px) + caption + mentions};
  \node[box, right=0.4cm of crop] (vlm) {Vision llama-server\\{\scriptsize Qwen3-VL-4B, started on first use}};
  \node[box, right=0.4cm of vlm] (fval) {Validate\\{\scriptsize schema; one retry}};
  \node[result, right=0.4cm of fval] (fpop) {Popup\\{\scriptsize what it shows, how to read it, takeaway}};
  \draw[arr] (boxsel) -- (crop); \draw[arr] (crop) -- (vlm); \draw[arr] (vlm) -- (fval); \draw[arr] (fval) -- (fpop);
  \draw[arr, dashed] (doc.south) |- (boxsel.west);
\end{tikzpicture}
\caption{The \aynaz{} pipeline. Text selections (top) and figure boxes (bottom) share the document model; both answers are schema-constrained, stream into the popup as they are generated and are validated, with one retry, before they are final. Everything runs on the reader's computer.}
\label{fig:pipeline}
\end{figure*}

\section{Related Work}

\paragraph{Reading interfaces for technical documents.}
ScholarPhi augments scientific papers with just-in-time, position-sensitive definitions of terms and symbols that a paper defines itself \citep{head2021scholarphi}.
The Semantic Reader project builds AI-powered interactive reading interfaces over existing PDFs \citep{lo2024semanticreader}, and Paper Plain helps healthcare consumers read medical papers with term definitions, plain-language section summaries and a question-guided index \citep{august2023paperplain}.
\aynaz{} shares their goal of help in place, without leaving the document.
It differs in generating the explanation for any selection with a local language model, including common words whose sense depends on the field rather than terms the paper defines, in covering figures, and in running entirely on the reader's computer.

\paragraph{Word sense and grounding.}
Choosing the sense a context intends is word sense disambiguation \citep{navigli2009wsd}.
\aynaz{} does not select from a sense inventory; it asks the model to explain the word as the document uses it, conditioned on passages retrieved from the same document, in the spirit of retrieval-augmented generation \citep{lewis2020rag} with a classical BM25 retriever \citep{robertson2009bm25}.

\paragraph{Structured output and small models.}
Constraining decoding to a grammar or schema makes a model's output machine-readable by construction \citep{willard2023outlines}; \aynaz{} uses llama.cpp's JSON-schema constrained decoding \citep{llamacpp}.
Qwen3 offers dense models from 0.6B parameters with a per-request switch between thinking and non-thinking modes \citep{yang2025qwen3}, and Qwen3-VL adds vision \citep{qwen2025qwen3vl}.

\paragraph{Figures.}
Scientific figure captioning \citep{hsu2021scicap} and chart question answering \citep{masry2022chartqa} study understanding figures as a task in itself; \aynaz{} instead explains a figure for a reader, grounded in its caption and the passages that mention it.

\paragraph{Model judges.}
Strong models are widely used as judges of open-ended answers, with known biases \citep{zheng2023judge}; we use one and report how stable its verdicts are (\S\ref{sec:judge}).

\section{System}
\label{sec:system}

Figure~\ref{fig:pipeline} shows the pipeline.
\aynaz{} is an Electron application; the renderer owns the document view and the popup, and the main process owns the inference servers, files and cache.

\paragraph{Document model.}
Every format is rendered into the same document object model (DOM) of the window: PDF through PDF.js text layers \citep{pdfjs}, DOCX through a DOCX renderer, Markdown and plain text directly, and EPUB, PPTX, XLSX, ODT and RTF through small converters into a common document model.
A reading-order text index walks that DOM and maps both ways between DOM positions and character offsets, so selection, context window, retrieval and locators (page, slide, sheet or chapter) work on one string, and the rest of the pipeline never branches on file type.

\paragraph{Context packet.}
For a selection $s$ \aynaz{} builds a packet $X(s)$ of about 2k tokens,
\begin{equation}
  X(s) = \big(T, O, F, \ell(s), s, W_2(s), R_3(s)\big),
  \label{eq:packet}
\end{equation}
where $T$ is the title, $O$ the opening 1{,}200 characters of the document, $F$ the document's field (inferred once per document by a short schema-constrained call), $\ell(s)$ the locator, $s$ the selection (at most 600 characters, beyond which the gist rather than a term is explained), $W_2(s)$ a window of $\pm$2 sentences around it capped at 1{,}500 characters, and $R_3(s)$ the three passages with the highest BM25 score \citep{robertson2009bm25} that do not overlap the window.
The passages are chunks $c$ of at most 1{,}000 characters (about 250 tokens) with a 150-character overlap, scored against the selection's tokens $Q$ (lower-cased letters and digits, stop words removed):
\begin{equation}
  \mathrm{BM25}(Q, c) = \sum_{t \in Q} \mathrm{idf}(t)\, \frac{f_{t,c}\,(k_1 + 1)}{f_{t,c} + K_c},
  \label{eq:bm25}
\end{equation}
with $K_c = k_1 (1 - b + b\, L_c / \bar{L})$ and $\mathrm{idf}(t) = \ln\!\big(1 + (M - n_t + 0.5)/(n_t + 0.5)\big)$, where $f_{t,c}$ is the count of token $t$ in chunk $c$, $L_c$ the chunk's length in tokens, $\bar{L}$ the mean length, $M$ the number of chunks and $n_t$ the number containing $t$; $k_1 = 1.2$ and $b = 0.75$.
All document text is fenced as data, and the instructions say text inside the fences is not to be followed.

\paragraph{Generation and validation.}
The packet goes to a local llama-server (llama.cpp build b11836) running the official Qwen3-4B checkpoint in llama.cpp's GGUF format with Q4\_K\_M quantisation (mostly 4-bit weights, about 5 bits per weight on average; 2.50\,GB) with an 8{,}192-token context, thinking off, temperature 0.2, a fixed seed of 42 and at most 700 output tokens.
Decoding is constrained to a JSON schema with \texttt{term}, \texttt{meaning}, \texttt{example}, a non-empty \texttt{diagram\_mermaid}, a \texttt{confidence} level and the pages used.
The meaning streams into the popup as it is generated.
The finished answer is then validated: the schema, word limits checked after decoding (60 words for the meaning, 80 for the example) and a parse of its Mermaid code \citep{mermaid}; a failure triggers one retry with the error message added to the conversation.
An answer is cached under the document's hash, the prompt version, the model file, the depth and the selection.
A \emph{Deeper} button re-asks with thinking on (512-token reasoning budget), a $\pm$4-sentence window and five passages (not evaluated in this paper).

\paragraph{Prompt rules.}
The explanation prompt asks the model to explain the selection as this document uses it, with two worked examples.
Its rules, summarised here rather than quoted, came out of development and the evaluation: (i) a short clause saying which other meaning is \emph{not} intended is added only for a plausible confusion with another field's sense, never as a contrast with an everyday thing (early versions invented contrasts, such as reading a metric's name as a colour); (ii) the local sentence decides the sense even when it departs from the document's overall topic, and a word that is part of a multi-word term is explained within that term; (iii) a definition given in the document is preferred to a general one; and (iv) a diagram is always required, and the prompt shows the Mermaid syntax it needs.
Section~\ref{sec:ablation} measures the revision that introduced (i) and (ii).

\paragraph{Figures.}
The reader draws a box over a figure; \aynaz{} captures that region (long side at most 1{,}024 pixels), finds caption candidates on and around the box edge (up to 800 characters) and the passages that mention the figure, and sends them with the image to a second llama-server running Qwen3-VL-4B with its multimodal projector.
That server starts on the first figure and stops after five minutes without use.
The answer follows a three-part schema (what it shows, how to read it, takeaway) and is validated like a text answer.

\paragraph{Outside the reader and privacy.}
A global hotkey explains text selected in other applications by copying the selection and restoring the clipboard; it has only the selection as context and says so (not evaluated in this paper).
The application makes no network requests except the one-time model download and a daily check of a small version file on the project's website.
The installer is about 180--225\,MB depending on the platform; the text model (2.50\,GB) and, for figures, the vision model with its projector (2.95\,GB) are downloaded on first use and checked by size and SHA-256.

\section{Evaluation Setup}
\label{sec:eval}

\paragraph{Corpus and cases.}
We built the evaluation from \SetPapers{} open-licence papers (CC~BY or CC0) in \SetFields{} fields, plus excerpts of some of them converted to DOCX, Markdown, plain text, PowerPoint and EPUB, \SetDocs{} documents in all.
The \SetCases{} cases (Table~\ref{tab:composition}) are of three kinds: \SetKindPolysemous{} \emph{polysemous} cases, a word with several meanings across fields such as \emph{kernel}, \emph{power}, \emph{spectrum} or \emph{resistance}; \SetKindTerm{} field terms, a term used mainly in one field such as \emph{SVM} (support vector machine); and \SetKindPassage{} longer passages, whose gist is scored rather than a sense.
The cases were chosen during development from terms that occur in the documents, favouring words with senses in several fields.
\SetMultiFieldTerms{} terms occur in the documents of two or more fields, which yields \MainPairN{} \emph{cross-field pairs}: the same term in two fields' documents, counted correct only when both explanations pick the right sense.
\SetNonPdf{} cases are on non-PDF formats.

\begin{table}[t]
\centering\small
\begin{filecontents*}{data/composition.csv}
field,polysemous,term,passage,total
Biology,12,3,1,16
CS/AI,15,8,1,24
Economics,11,3,1,15
Elec. eng.,14,3,1,18
Materials,2,2,1,5
Medicine,10,4,1,15
Physics,22,4,1,27
Statistics,7,3,1,11
All fields,93,30,8,131
\end{filecontents*}
\pgfplotstabletypeset[col sep=comma, string type,
  columns={field,polysemous,term,passage,total},
  columns/field/.style={column name=Field, column type=l},
  columns/polysemous/.style={column name=Polys., column type=r},
  columns/term/.style={column name=Term, column type=r},
  columns/passage/.style={column name=Passage, column type=r},
  columns/total/.style={column name=Total, column type=r},
  every head row/.style={before row=\toprule, after row=\midrule},
  every last row/.style={before row=\midrule, after row=\bottomrule}]{data/composition.csv}
\caption{Evaluation cases by field and kind. By format: PDF \FmtPdf, DOCX \FmtDocx, Markdown \FmtMd, EPUB \FmtEpub, PowerPoint \FmtPptx, plain text \FmtTxt.}
\label{tab:composition}
\end{table}

\paragraph{Reference answers and judge.}
For each case a judge model, Qwen3-30B-A3B (Q4\_K\_M) on a second llama-server, drafted a reference: the intended sense, a gist and realistic wrong senses.
Every draft was then reviewed against the document text by an AI assistant, at the author's request; in the first 120 cases this corrected 12 senses and 15 gists.
The same judge then scores each answer against its reference and the document passages: an overall score $q_i$ from 1 to 5, a sense verdict $\sigma_i \in \{0, 1\}$, a category for the ``not the \ldots'' clause (\emph{none}, \emph{apt}, \emph{silly}, or \emph{missing} where one was needed), and a 1--5 score for the diagram's logic (Appendix~\ref{app:judge}).
Its rubric defines the overall score as 5: right sense, accurate and grounded in the document; 4: right sense with a minor omission or slight vagueness; 3: right sense but generic, or one small factual slip; 2: partly wrong or confusing; 1: wrong sense or wrong facts.

\paragraph{Metrics.}
With $\mathcal{P}$ the polysemous cases, $\Pi$ the cross-field pairs and $N$ the number of cases, \emph{sense accuracy} (SA), \emph{pair accuracy} (PA) and \emph{correct in context} (CIC) are
\begin{gather}
  \mathrm{SA} = \frac{1}{|\mathcal{P}|} \sum_{i \in \mathcal{P}} \sigma_i, \qquad
  \mathrm{PA} = \frac{1}{|\Pi|} \sum_{(i,j) \in \Pi} \sigma_i \sigma_j, \nonumber \displaybreak[0] \\
  \mathrm{CIC} = \frac{1}{N} \sum_{i=1}^{N} \mathbb{1}[q_i \ge 4].
  \label{eq:metrics}
\end{gather}
\emph{Schema validity} and the \emph{diagram render rate} are measured by the application itself.
For a proportion $\hat{p} = k/n$ we give the Wilson 95\,\% confidence interval (CI) \citep{wilson1927},
\begin{equation}
  \frac{\hat{p} + \frac{z^2}{2n} \pm z \sqrt{\frac{\hat{p}(1 - \hat{p})}{n} + \frac{z^2}{4n^2}}}{1 + \frac{z^2}{n}}, \qquad z = 1.96.
  \label{eq:wilson}
\end{equation}
Cross-field pairs share cases, so their outcomes are not independent and we give no interval for PA.
Latency is measured in the renderer from the start of generation, after the context packet has been built, to the first visible words of the meaning (\emph{first text}) and to the validated answer (\emph{full}).

\paragraph{Procedure.}
A test runner drives the real interface path for every case: it opens the document, finds the selection's occurrence, sets the DOM selection and calls the same function a mouse selection calls; the answer cache is bypassed.
All results were measured on one desktop computer (Windows~11, AMD Ryzen~9 9900X3D, 61\,GB RAM, two NVIDIA RTX 5070 Ti 16\,GB GPUs).
The word evaluation used llama.cpp b11836's CUDA build; the released application uses the Vulkan build of the same version, compared in \S\ref{sec:engine}.
The same answers were judged in two server configurations: Judge~A is the later one, with each server on one GPU, and Judge~B the earlier one, with the servers split over both GPUs (\S\ref{sec:judge}).

\section{Results}
\label{sec:results}

\subsection{Words}
\label{sec:words}

\begin{table}[t]
\centering\small
\setlength{\tabcolsep}{4pt}
\begin{tabular}{@{}lccc@{}}
\toprule
Metric & Judge A & Judge B & Target \\
\midrule
Sense accuracy & \pct{\MainSense} & \pct{\AltSense} & $\ge$95\,\% \\
\quad 95\,\% CI & \scriptsize\ci{\MainSenseLo}{\MainSenseHi} & \scriptsize\ci{\AltSenseLo}{\AltSenseHi} & \\
Pair accuracy (\MainPairN{} pairs) & \pct{\MainPair} & \pct{\AltPair} & -- \\
Correct in context & \pct{\MainCorrect} & \pct{\AltCorrect} & $\ge$90\,\% \\
\quad 95\,\% CI & \scriptsize\ci{\MainCorrectLo}{\MainCorrectHi} & \scriptsize\ci{\AltCorrectLo}{\AltCorrectHi} & \\
Diagram logic, mean (1--5) & \MainLogic & \AltLogic & -- \\
``Not the'' clause: silly & \MainClauseSilly & \AltClauseSilly & -- \\
\midrule
Schema valid & \multicolumn{2}{c}{\pct{\MainSchema}} & $\ge$99.5\,\% \\
Diagram renders & \multicolumn{2}{c}{\pct{\MainDiagram}} & $\ge$95\,\% \\
First text, median / p90 & \multicolumn{2}{c}{\MainFirstMed{} / \MainFirstNinety{}\,ms} & $\le$1\,s \\
Full answer, median / p90 & \multicolumn{2}{c}{\MainFullMed{} / \MainFullNinety{}\,ms} & $\le$3\,s \\
\bottomrule
\end{tabular}
\caption{Results on the \SetCases{} cases (prompt v2, Qwen3-4B Q4\_K\_M, llama.cpp CUDA build). The answers are identical under both columns; A and B are the same judge model on two server configurations (\S\ref{sec:judge}). System-side metrics do not depend on the judge. p90: 90th percentile.}
\label{tab:main}
\end{table}

Table~\ref{tab:main} gives the results.
\aynaz{} picks the document's sense in \MainSenseK{} of \MainSenseN{} polysemous cases (\pct{\MainSense}) and both senses of \MainPairK{} of \MainPairN{} cross-field pairs; \pct{\MainCorrect} of all answers are judged correct in context.
Every answer satisfied the schema, \pct{\MainDiagram} of diagrams rendered, and \MainRetries{} case needed the retry.
The point estimates meet every design target set before the evaluation (right column); the lower end of the sense-accuracy interval (\pct{\MainSenseLo}) lies below the 95\,\% target, so this sample does not establish the target for new documents.

\begin{figure}[t]
\centering
\begin{tikzpicture}
\begin{axis}[width=\columnwidth, height=4.8cm,
  ymin=0, ymax=105, ytick={0,20,40,60,80,100}, ylabel={Sense accuracy (\%)}, ylabel style={font=\small},
  xtick=data, xticklabels from table={data/field-sense.csv}{label}, table/col sep=comma,
  x tick label style={rotate=30, anchor=north east, font=\scriptsize, inner sep=1pt}, yticklabel style={font=\scriptsize},
  enlarge x limits=0.08, axis lines*=left, ymajorgrids, grid style={black!10}]
\addplot[only marks, mark=*, mark size=2.2pt, color=accent, error bars/.cd, y dir=both, y explicit, error bar style={black!60, thick}]
  table[col sep=comma, x expr=\coordindex, y=acc, y error plus=ep, y error minus=em]{data/field-sense.csv};
\end{axis}
\end{tikzpicture}
\caption{Sense accuracy per field (dots) with Wilson 95\,\% intervals; each label gives the field's number of polysemous cases.}
\label{fig:fields}
\end{figure}
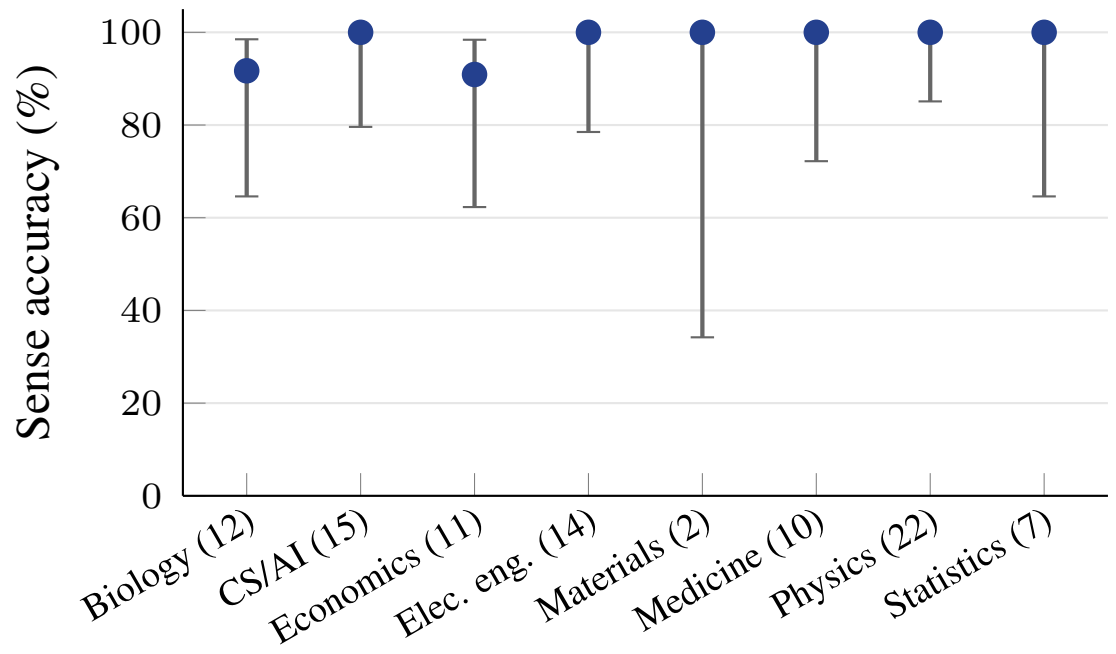

Figure~\ref{fig:fields} breaks sense accuracy down by field.
The intervals are wide because each field has 2 to 22 polysemous cases; the two errors fall in biology and economics.
In biology, \emph{cell} in a study of host cells infected by \emph{Chlamydia} was explained as the pathogen's own developmental forms; in economics, a \emph{positive} shock was read as a beneficial one rather than a price-raising one.
Three further answers had the right sense but a factual slip and scored 3, for example confusing a current density with a power in a solar-cell paper.

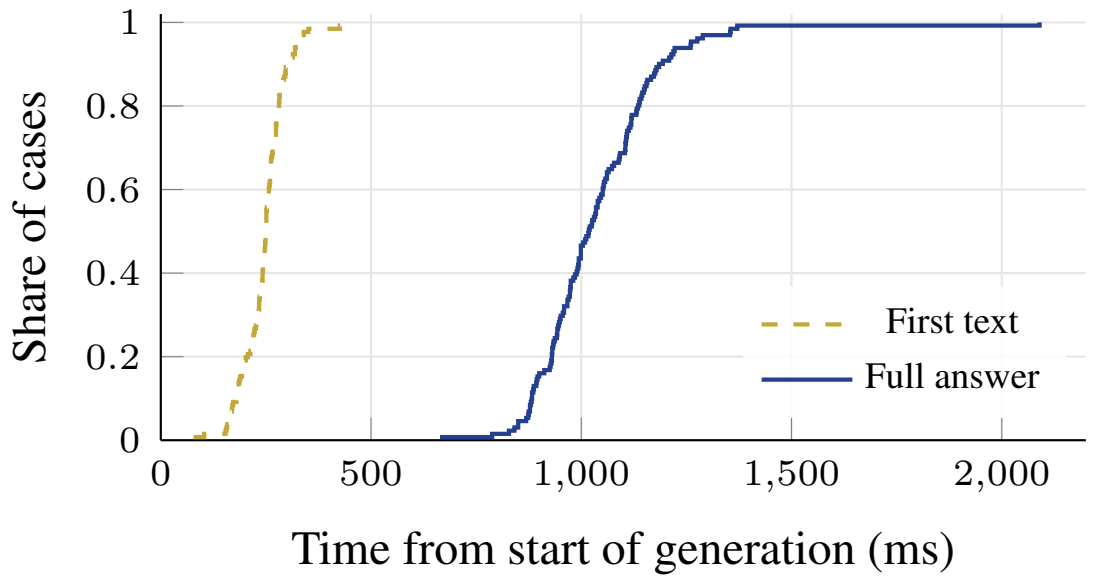
\begin{figure}[t]
\centering
\begin{tikzpicture}
\begin{axis}[width=\columnwidth, height=4.4cm, xmin=0, xmax=2200, ymin=0, ymax=1.02,
  xlabel={Time from start of generation (ms)}, ylabel={Share of cases}, label style={font=\small}, tick label style={font=\scriptsize},
  legend style={font=\scriptsize, at={(0.98,0.05)}, anchor=south east, draw=none, fill=white, fill opacity=0.8, text opacity=1},
  axis lines*=left, grid=major, grid style={black!10}]
\addplot[const plot, thick, dashed, hl!85!black] table[col sep=comma, x=ms, y=frac]{data/latency-first.csv};
\addlegendentry{First text}
\addplot[const plot, thick, accent] table[col sep=comma, x=ms, y=frac]{data/latency-full.csv};
\addlegendentry{Full answer}
\end{axis}
\end{tikzpicture}
\caption{Cumulative latency over the \SetCases{} cases on an RTX 5070 Ti.}
\label{fig:latency}
\end{figure}

\paragraph{Latency.}
Figure~\ref{fig:latency} shows the latency distribution.
The first words of the meaning appear after a median of \MainFirstMed{}\,ms (90th percentile \MainFirstNinety{}\,ms) and the validated answer after \MainFullMed{}\,ms (90th percentile \MainFullNinety{}\,ms, maximum \MainFullMax{}\,ms).
During development, one warm call on the same computer with the GPU libraries removed took 7.9\,s for the full answer and 450\,ms to the first text; this is a single measurement with an earlier prompt, not an evaluation of CPU-only use.

\subsection{How stable is the judge?}
\label{sec:judge}

The two columns of Table~\ref{tab:main} come from two evaluation runs made on the same day with the same prompt and model.
Between them the inference servers moved from two GPUs to one, which changed the judge's server configuration as well.
The application's \JudgeSameAnswers{} answers are byte-for-byte identical in both runs: with a fixed seed (42), temperature 0.2 and the same engine build, the answers did not change although the GPU placement had.
The judge's verdicts on those identical answers are not: \JudgeVerdictsChanged{} of \SetCases{} verdicts changed, \JudgeScoreChanged{} overall scores moved (each by at most \JudgeScoreMaxStep{} point), \JudgeClauseChanged{} clause categories changed and \JudgeSenseFlips{} sense verdicts flipped.
Under an unchanged configuration the judge is repeatable: three runs made before the change (the configuration of Judge~B) gave identical verdicts on all \JudgeSameConfigAgree{} cases they share.
Sense accuracy therefore reads \pct{\AltSense} or \pct{\MainSense} for the same answers, depending only on the judge's server.
Both flipped cases are borderline (a \emph{positive} shock, and \emph{Hilbert space} explained correctly but without the document's focus on a reproducing kernel Hilbert space).
Under Judge~A both still received a score of 4 although the judge marked their sense wrong; its rubric reserves a 4 for the right sense, so these two verdicts are inconsistent.
Counting them as incorrect gives \MainCorrectStrictK{} of \MainCorrectStrictN{} (\pct{\MainCorrectStrict}, 95\,\% CI \ci{\MainCorrectStrictLo}{\MainCorrectStrictHi}) correct in context; the point estimate is still above the 90\,\% target, but the interval's lower end is below it.
We report the lower figures as the headline and read differences of two or three cases between configurations as within the judge's own variation.

\subsection{Prompt ablation}
\label{sec:ablation}

\begin{table}[t]
\centering\small
\setlength{\tabcolsep}{3.5pt}
\begin{tabular}{@{}lcccc@{}}
\toprule
Prompt & Sense & Pairs (116) & Correct & Silly \\
\midrule
v1 & \pct{\VoneSense} & \pct{\VonePair} & \pct{\VoneCorrect} & \VoneClauseSilly \\
v2 (shipped) & \pct{\VtwoSense} & \pct{\VtwoPair} & \pct{\VtwoCorrect} & \VtwoClauseSilly \\
v3 & \pct{\VthreeSense} & \pct{\VthreePair} & \pct{\VthreeCorrect} & \VthreeClauseSilly \\
\bottomrule
\end{tabular}
\caption{Prompt versions on the first 120 cases (84 polysemous). v2 restricts the ``not the \ldots'' clause, replaces the one worked example with two and adds rule (ii); v3 adds a rule to follow the sentence when retrieved passages disagree. \emph{Silly}: clauses naming a contrast that does not exist.}
\label{tab:ablation}
\end{table}

Table~\ref{tab:ablation} compares three prompt versions on the first 120 cases.
The clearest effect is on the ``not the \ldots'' clause.
v1 already asked for the clause only when the word has a common other meaning, but it counted everyday meanings as candidates and its worked example contrasted a term with an everyday object; the model produced invented contrasts in \VoneClauseSilly{} cases.
v2 restricted the clause to plausible confusions with another field, ruled out everyday contrasts and replaced the example; silly clauses fell to \VtwoClauseSilly{}.
All three versions were judged under the same configuration, where verdicts are repeatable, so these differences come from the answers.
Pair accuracy rose from \VonePairK{} to \VtwoPairK{} of 116 pairs, more than the two pairs that a configuration change alone moved in \S\ref{sec:judge}.
Because v2 changed several things at once, the table measures the revision as a whole, not one rule.
The sense-accuracy gain of two cases is no larger than the two sense verdicts that a configuration change alone flipped in \S\ref{sec:judge}, and it lies well inside the sampling interval, so we do not claim it.
v3 brought no improvement and was not adopted.
v2 was tuned on these same cases, so its numbers are optimistic for unseen documents.

\subsection{Engine and quantisation}
\label{sec:engine}

\begin{table*}[t]
\centering\small
\setlength{\tabcolsep}{4.5pt}
\begin{tabular}{@{}lrrcccccc@{}}
\toprule
Engine + model & Engine & Model & First text / full (ms) & Sense & Pairs & Correct & Diagrams & Silly \\
\midrule
CUDA + Q4\_K\_M\textsuperscript{a} & 726\,MB & 2.50\,GB & 245 / 1008 & \pct{\MainSense} & \pct{\MainPair} & \pct{\MainCorrect} & \MainDiagramK/\MainDiagramN & \MainClauseSilly \\
Vulkan + Q4\_K\_M (shipped) & 100\,MB & 2.50\,GB & 298 / 1072 & 97.8\,\% & 97.0\,\% & 95.4\,\% & 130/131 & 3 \\
Vulkan + IQ4\_XS & 100\,MB & 2.27\,GB & 309 / 1073 & 97.8\,\% & 95.6\,\% & 96.9\,\% & 120/131 & 3 \\
Vulkan + IQ3\_M & 100\,MB & 1.96\,GB & 314 / 1077 & 95.7\,\% & 92.6\,\% & 94.7\,\% & 126/131 & 14 \\
\bottomrule
\end{tabular}
\caption{Engine and quantisation trade-off, from the development record (aggregates only; each configuration judged once; the judge's own configuration was not recorded). Latencies come from separate speed runs recorded with the same table, so they differ slightly from Table~\ref{tab:main}. \textsuperscript{a}The quality columns of this row are the Judge~A run of Table~\ref{tab:main}. Given \S\ref{sec:judge}, differences of one or two cases are not meaningful.}
\label{tab:quant}
\end{table*}

The released application uses llama.cpp's Vulkan build on Windows and Linux, which is 100\,MB instead of 726\,MB for the CUDA build and also runs on AMD and Intel GPUs; macOS uses Metal.
All measurements in this paper are on Windows; the Linux and macOS builds were not measured.
Table~\ref{tab:quant} shows the development measurements behind the choice of engine.
On this NVIDIA GPU the Vulkan build was about 6\,\% slower than CUDA to the full answer (and 298 versus 245\,ms to the first text), with the same sense and pair accuracy in this single judging.
The smaller IQ4\_XS quantisation saved 0.23\,GB but its diagram render rate fell below the 95\,\% target, through a repeated Mermaid syntax slip.
IQ3\_M lost six cross-field pairs and multiplied silly clauses, both beyond the judge's variation; its sense-accuracy drop of two cases is within it.
Q4\_K\_M was kept.

\subsection{Figures}

\begin{table}[t]
\centering\small
\begin{tabular}{@{}lcc@{}}
\toprule
Metric & Value & 95\,\% CI \\
\midrule
Correct in context ($\ge$4) & \FigCorrectK/\FigCorrectN & \ci{\FigCorrectLo}{\FigCorrectHi} \\
Grounded in the document's terms & \FigGroundedK/\FigGroundedN & \ci{\FigGroundedLo}{\FigGroundedHi} \\
Invented details & \FigInventedK/\FigInventedN & \ci{\FigInventedLo}{\FigInventedHi} \\
Schema valid & \FigSchemaK/\FigSchemaN & \\
Full answer, median / max & \multicolumn{2}{c}{\FigMedMs{} / \FigMaxMs{}\,ms} \\
\bottomrule
\end{tabular}
\caption{Figure explanations on \FigCases{} figures and tables from \FigDocs{} papers (Qwen3-VL-4B). The maximum includes starting the vision server.}
\label{tab:figures}
\end{table}

\begin{figure}[t]
\centering
\includegraphics[width=0.92\columnwidth]{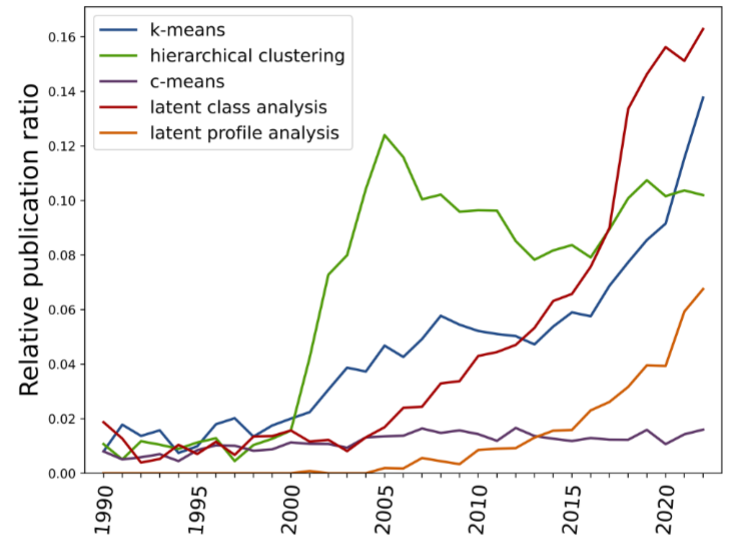}\\[3pt]
\fbox{\parbox{0.9\columnwidth}{\footnotesize
\textbf{What it shows:} Relative popularity of different statistical methods in publications, measured by citation frequency, from 1990 to 2020.\\[2pt]
\textbf{How to read it:} Vertical axis is relative publication ratio; horizontal axis is year. Each line represents a method: k-means, hierarchical clustering, c-means, latent class analysis, latent profile analysis.\\[2pt]
\textbf{Takeaway:} Latent class and latent profile analyses show rapid growth since 2015, while hierarchical clustering peaks around 2005.}}
\caption{One of the \FigCases{} figure cases and \aynaz{}'s answer, verbatim (950\,ms). The judge scored it 5, but the answer has two slips: the plot runs to 2022, not 2020, and its caption measures papers that mention each method relative to papers that mention the \emph{t}-test, not citation frequency. Figure from Dalmaijer, arXiv:2309.00866, CC~BY~4.0.}
\label{fig:figure-example}
\end{figure}

On \FigCases{} figures and tables from \FigDocs{} papers, all answers were judged correct in context, \FigScoreFive{} with the top score, and one invented a detail (a colour mix-up); Table~\ref{tab:figures} gives the intervals, which are wide at this sample size, and Appendix~\ref{app:figures} lists every case.
Figure~\ref{fig:figure-example} shows an example.
The figure path was tuned on these cases: a first pass during development was correct in 92.9\,\% of cases with invented details in 14.3\,\%, and reading captions up to 800 characters, including captions on the edge of the box, and giving the judge the figure's passages brought it to the results above.
A revised figure prompt tied with the first and was not adopted.

\subsection{Qualitative examples}

Three of the six \emph{resistance} cases show the cross-field behaviour.
In the clinical study the answer begins ``The ability of bacteria to survive antibiotic treatment, often due to genetic changes\ldots''; in the materials paper ``The ability of a material to withstand mechanical deformation without breaking''; in the electrical-engineering paper ``A component in electrical circuits that opposes current flow\ldots'', followed by an apt clause, ``(Not the everyday object or material.)''.
All three were scored 5, although the last describes a resistor rather than resistance; the judge did not penalise this imprecision.
Two answers scored 1: the biology \emph{cell} answer of \S\ref{sec:words} and a field-term case outside the sense count, which explained \emph{SVM} correctly in general but claimed the paper's SVM was quantum-enhanced, when the document compares a quantum model against a classical SVM.

\section{Limitations}

\begin{itemize}
  \item \textbf{Model judge, AI-reviewed references.} Every quality number comes from a 30B model of the same family as the system, and the references were drafted by that model and reviewed by an AI assistant, not by human domain experts. The judge also scored a loose answer (``a component \ldots'') 5, and its verdicts moved between server configurations (\S\ref{sec:judge}).
  \item \textbf{Small, tuned evaluation.} \SetCases{} text cases and \FigCases{} figures; prompt v2 and the figure fixes were developed on the same cases that are reported, so the numbers are optimistic for new documents.
  \item \textbf{No baselines.} We did not measure the same model without the document context, without retrieval, or against other systems, so the contribution of each part of the packet is not isolated.
  \item \textbf{Figures.} The figure references were written by the AI assistant from the image crops, and the figure judge does not see the image; it judges the answer against the reference, caption and passages.
  \item \textbf{One computer, English only, no user study.} Latency was measured on one high-end consumer GPU; the CPU-only figure is a single call. The evaluation covers English documents only, and we have not measured whether the explanations help readers understand.
\end{itemize}

\section{Conclusion}

\aynaz{} shows that a 4-bit (Q4\_K\_M) quantised 4B model on the reader's own computer can explain a selected word as the document uses it, with the right sense in about 98\,\% of the polysemous cases we tested, the first words in about a quarter of a second and the full answer in about a second on one consumer GPU, and that a 4B vision model gave answers judged correct in context for all \FigCases{} figures and tables we tested.
It does so with a context packet that includes retrieval, schema-constrained decoding and a single validated retry; we have not yet measured how much each of these contributes.
The evaluation also shows that a model judge's verdicts can move more than the differences one hopes to measure.
The next steps are human evaluation of the references and the judge, baselines without context and without retrieval, a user study with readers, and documents beyond English.

\section*{Availability}
\aynaz{} is free to download for Windows, macOS (Apple silicon) and Linux at \url{https://azharul-islam007.github.io/aynaz/}.
The source code, prompts and evaluation data are not publicly released.

\section*{Acknowledgements}
The evaluation documents are open-access papers under CC~BY or CC0 licences.
An AI assistant was used in developing the software and in preparing this manuscript; the author reviewed all content and takes full responsibility for it.

\bibliography{references}

\appendix

\section{Figure cases}
\label{app:figures}

\begin{table}[H]
\centering\scriptsize
\begin{filecontents*}{data/figure-cases.csv}
field,label,score,invented,ms
Biology,Figure 1,5,0,4420
Biology,Figure 2,5,0,958
Biology,Figure 1,5,0,1169
Biology,Figure 4,5,0,904
CS/AI,Figure 1,5,0,867
CS/AI,Figure 6,5,0,735
CS/AI,Figure 2,5,0,908
CS/AI,Table 1,5,0,747
CS/AI,Figure 2,5,0,871
CS/AI,Table 1,5,0,857
CS/AI,Figure 3,5,0,813
Economics,Figure 1,5,0,844
Economics,Figure 3,4,1,766
Economics,Table 1,5,0,776
Elec. eng.,Figure 3,5,0,907
Elec. eng.,Table 1,5,0,776
Elec. eng.,Figure 2,5,0,1039
Elec. eng.,Figure 5,5,0,864
Materials,Figure 1,5,0,882
Materials,Figure 3,5,0,862
Medicine,Table 1,5,0,819
Medicine,Table 2,5,0,747
Medicine,Table 1,4,0,986
Medicine,Table 3,5,0,1062
Physics,Figure 1,5,0,1034
Physics,Figure 1,5,0,923
Statistics,Figure 1,5,0,950
Statistics,Table 1,5,0,936
\end{filecontents*}
\pgfplotstabletypeset[col sep=comma, string type,
  columns={field,label,score,invented,ms},
  columns/field/.style={column name=Field, column type=l},
  columns/label/.style={column name=Figure, column type=l},
  columns/score/.style={column name=Score, column type=c},
  columns/invented/.style={column name=Invented, column type=c},
  columns/ms/.style={column name=Full answer (ms), column type=r},
  every head row/.style={before row=\toprule, after row=\midrule},
  every last row/.style={after row=\bottomrule}]{data/figure-cases.csv}
\caption{All \FigCases{} figure cases: judge score (1--5), whether a detail was invented (1) and the time to the full answer. The first case's time includes starting the vision server.}
\end{table}

\section{Judge outputs}
\label{app:judge}
For each text case the judge returns an overall score (1--5) for whether the explanation is correct and helpful for this document, a boolean for the sense, the category of the ``not the \ldots'' clause, and a 1--5 score for whether the diagram's logic is right (a diagram can render and still be wrong).
For each figure case it returns a score (1--5), whether the answer uses the document's terms (grounded) and whether it invents numbers or details that are not in the reference, the caption, the text inside the box or the passages (the judge does not see the image).

\end{document}